\pdfoutput=1
\documentclass{article}

\PassOptionsToPackage{numbers, compress}{natbib}
\usepackage[preprint]{neurips_distshift_2021}

\usepackage[utf8]{inputenc} 
\usepackage[T1]{fontenc}    
\usepackage{hyperref}       
\usepackage{url}            
\usepackage{booktabs}       
\usepackage{amsfonts}       
\usepackage{nicefrac}       
\usepackage{microtype}      
\usepackage{xcolor}         
\usepackage[pdftex]{graphicx}
\usepackage{amssymb}
\usepackage{amsmath}
\usepackage{float}
\usepackage{pdflscape}
\usepackage{tabulary}
\usepackage{url}
\usepackage{longtable}
\usepackage{subfig}
\usepackage{array}
\usepackage{adjustbox}
\usepackage{tabularx}

\title{An Empirical Investigation of Model-to-Model Distribution Shifts in Trained Convolutional Filters}

\author{%
  Paul Gavrikov$^{1}$ \qquad Janis Keuper$^{1,2}$\\
  $^{1}$Institute for Machine Learning and Analytics (IMLA), Offenburg University, Germany \\
  $^{2}$Fraunhofer ITWM, Kaiserslautern, Germany\\
  \texttt{\{paul.gavrikov, janis.keuper\}@hs-offenburg.de} \\
}

\begin{document}

\maketitle

\begin{abstract}
We present first empirical results from our ongoing investigation of distribution shifts in image data used for various computer vision tasks. Instead of analyzing the original training and test data, we propose to study shifts in the learned weights of trained models. In this work, we focus on the properties of the distributions of dominantly used $3\times 3$ convolution filter kernels. We collected and publicly provide a data set with over half a billion filters from hundreds of trained CNNs, using a wide range of data sets, architectures, and vision tasks. Our analysis shows interesting distribution shifts (or the lack thereof) between trained filters along different axes of meta-parameters, like data type, task, architecture, or layer depth. We argue, that the observed properties are a valuable source for further investigation into a better understanding of the impact of shifts in the input data to the generalization abilities of CNN models and novel methods for more robust transfer-learning in this domain.\\
Data available at: \url{https://github.com/paulgavrikov/CNN-Filter-DB/}.
\end{abstract}

\section{Introduction}
\vspace{-0.3cm}
Despite their overwhelming success in the application to various vision tasks, the practical deployment of convolutional neural networks (CNNs) is still suffering from several inherent drawbacks. Two prominent examples are I) the dependence on very large amounts of annotated training data \cite{sun2017revisiting}, which is not available for all target domains and is expensive to generate; and II) still widely unsolved problems with the robustness and generalization abilities of CNNs \cite{akhtar2018threat} towards shifts of the input data distributions. One can argue that both problems are strongly related, since a common practical solution to I) is the fine-tuning \cite{tajbakhsh2016convolutional} of pre-trained models by small data sets from the actual target domain. This results in the challenge to find suitable pre-trained models based on data distributions that are "as close as possible" to the target distributions. Hence, both cases (I+II) imply the need to model and observe distribution shifts in the contexts of CNNs.\\
In this paper, we propose not to investigate these shifts in the input (image) domain, but rather in the weight distributions of the CNNs themselves. We argue that e.g. the distributions of trained convolutional filters in a CNN, which implicitly reflect the sub-distributions of the input image data which are actually utilized by a specific model, are more suitable and easier accessible representations for this task.     
\vspace{-0.3cm}
\section{Methods}
\vspace{-0.3cm}
\textbf{Data.} We collected a total of \textbf{391 publicly available CNN models} pre-trained for various visual tasks, 
recorded meta-data for each model, and manually categorized the training data into visually distinctive groups (data type) like \textit{natural scenes, medical ct, seismic,} or \textit{astronomy} for example.
All models were trained with full 32-bit precision but may have been trained with variously scaled inputs. The dominant subset is formed by image classification models trained on \textit{ImageNet1k} \cite{imagenet} (264 models). 
We extracted all trained convolution filters to get a heterogeneous and diverse representation. Hereby, only the widely used filters with a kernel size of $3\times 3$ were taken into account. Filters were only extracted from regular convolution layers; Transposed
convolution layers were not included. A total of \textbf{524,563,289 filters} from \textbf{13,015 layers} is used for the following study.


\textbf{Structure analysis.} We apply a full-rank PCA transformation to understand the underlying structure of the filters. A linear combination of principal components $v_{i}$ weighted by the coefficients $c_{i}$ and a bias $b_{i}$ then describes each filter: $f=\sum_{i}c_{i}v_{i}+b_{i}$ .
%
%
Figure \ref{fig:filter_basis} shows the principal components computed on the complete data set and various sub-sets. \\

\textbf{Measuring distribution shifts.} The divergence between two distributions is measured by the symmetric, non-negative variant of Kullback-Leibler \cite{kullback1951}. The shift $D$ of two filter sets is then defined by the sum of the divergence of the coefficient distributions $P_{i}, Q_{i}$ along every principal component index $i$. The sum is weighted by the ratio of variance $q_{i}$ explained by the $i$-th principal component. 
\begin{equation}
    \begin{split}
        \displaystyle D(P\parallel Q)&=\sum_{i}q_{i}\sum_{x\in {\mathcal {X}}}P_{i}(x)\log \left({\frac {P_{i}(x)}{Q_{i}(x)}}\right) + Q_{i}(x)\log \left({\frac {Q_{i}(x)}{P_{i}(x)}}\right)
        \end{split}
\end{equation}
\noindent To avoid undefined expressions, all probability distributions $F$ are set to hold $\forall x\in {\mathcal {X}}: F(x)\geq\epsilon$.

\vspace{-0.2cm}
\section{Empirical evaluation}
\vspace{-0.3cm}
\subsection{Comparison of filter structures}
\begin{figure}[H]
    \centering
    \begin{minipage}{.49\textwidth}
    \centering
        \begin{tabular}{m{0.94\textwidth}}
        (a) \adjustbox{valign=c}{\includegraphics[width=0.925\linewidth]{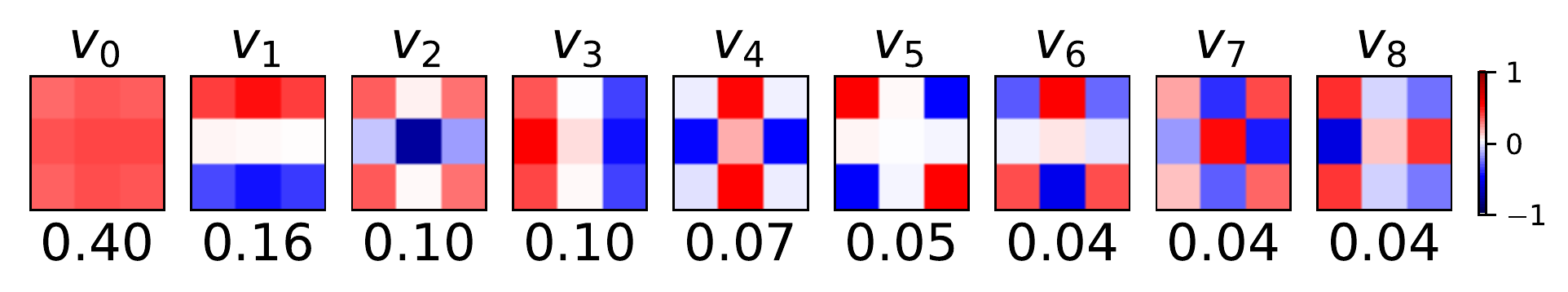}} \\ 
        (b) \adjustbox{valign=c}{\includegraphics[width=0.925\linewidth]{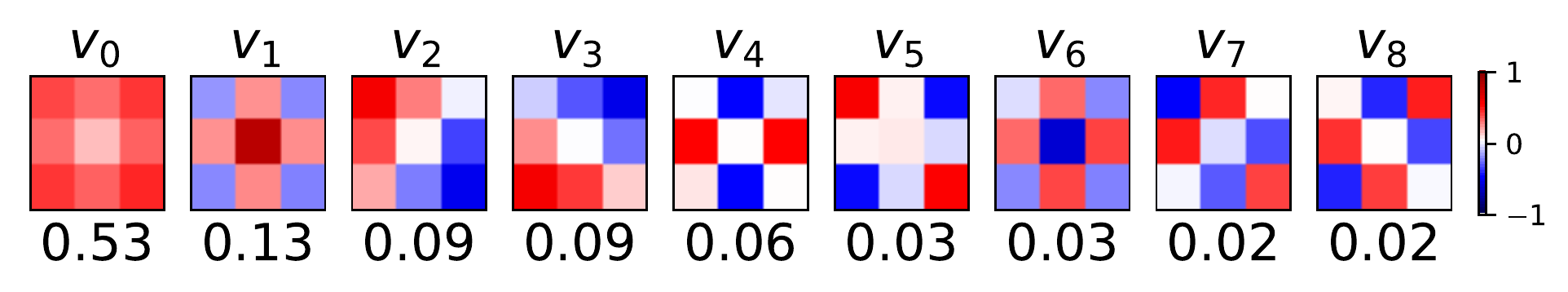}}\\
        (c) \adjustbox{valign=c}{\includegraphics[width=0.925\linewidth]{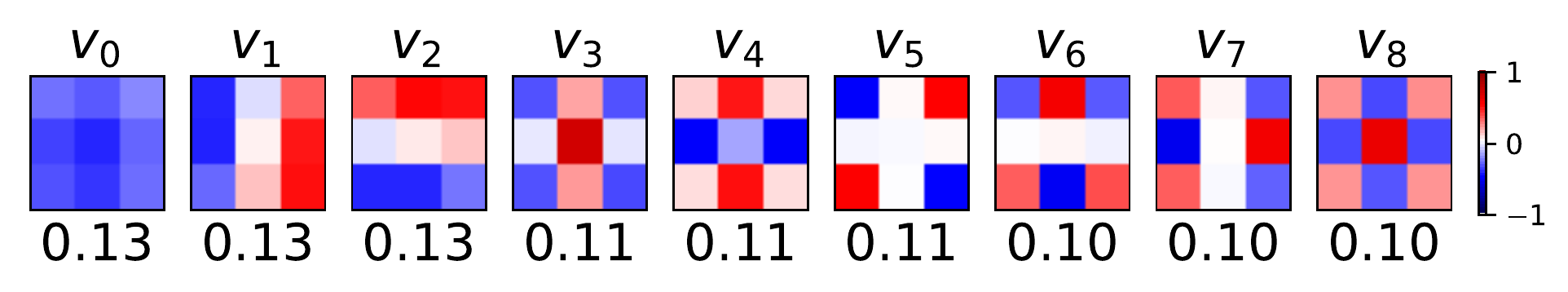}}\\
        (d) \adjustbox{valign=c}{\includegraphics[width=0.925\linewidth]{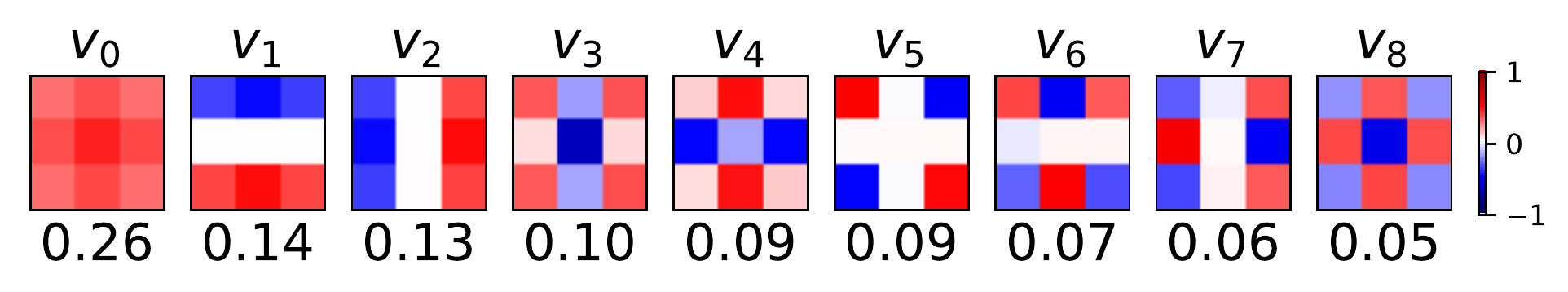}}
        \end{tabular}
        \vspace{0.5cm}
    \end{minipage} %
    \begin{minipage}{.49\textwidth}
        \centering
        
        \includegraphics[width=0.95\textwidth]{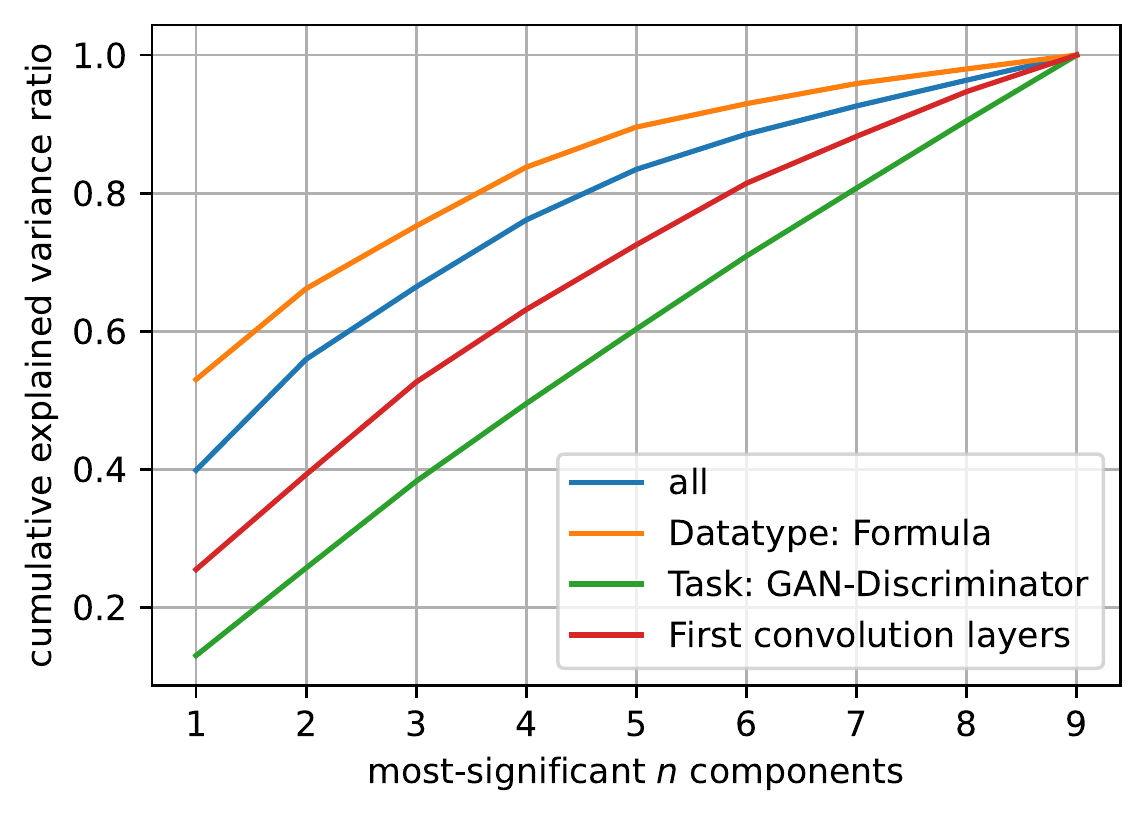}
    \end{minipage}
    
    \caption{Principal components $v_i$ and (cumulative) explained variance ratio per component for filters from (a) full dataset, (b) models trained on \textit{formula} data, (c) GAN discriminators, (d) first convolution layers. More examples in \autoref{sec:appendix_pca}.}
    \label{fig:filter_basis}
\end{figure}

\begin{figure}[H]
  \centering
  \includegraphics[width=\linewidth]{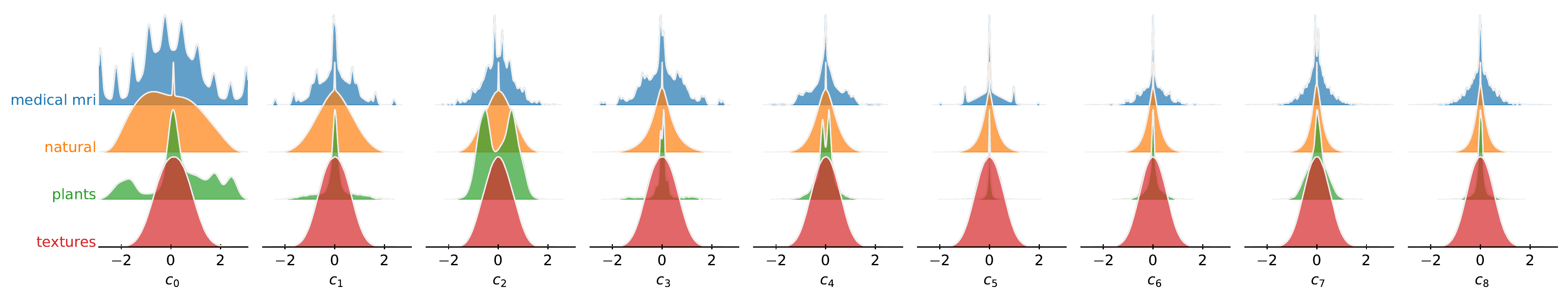}
  \caption{Coefficient distributions along the principal components for selected data types. Full overview in \autoref{sec:appendix_ridge_plots}.}
  \label{fig:ridge_selected}
\end{figure}

\vspace{-0.3cm}
In a first series of experiments, we analyze only the structure of $3\times 3$ filters, neglecting their actual numerical weight in the trained models. We scale all filters by their absolute maximum weight and perform a PCA. Figure \ref{fig:filter_basis} shows some qualitative examples of principal components, split by several meta-data dimensions. Figure \ref{fig:ridge_selected} depicts the shifts between distributions of PCA coefficients along models trained on different data types. We hypothesize that the spiky kernel density estimates (KDEs) are caused by "degenerated" layers e.g. if a model was too complex/deep for a given dataset (e.g. \textit{ResNet-101} \cite{he2015deep} on \textit{CIFAR-10} \cite{cifar10}).
\vspace{-0.2cm}

\vspace{-0.2cm}

\vspace{-0.3cm}
\subsubsection{Observable shifts along filters sub-sets}
\begin{figure}[H]
  \centering
  \includegraphics[width=\linewidth]{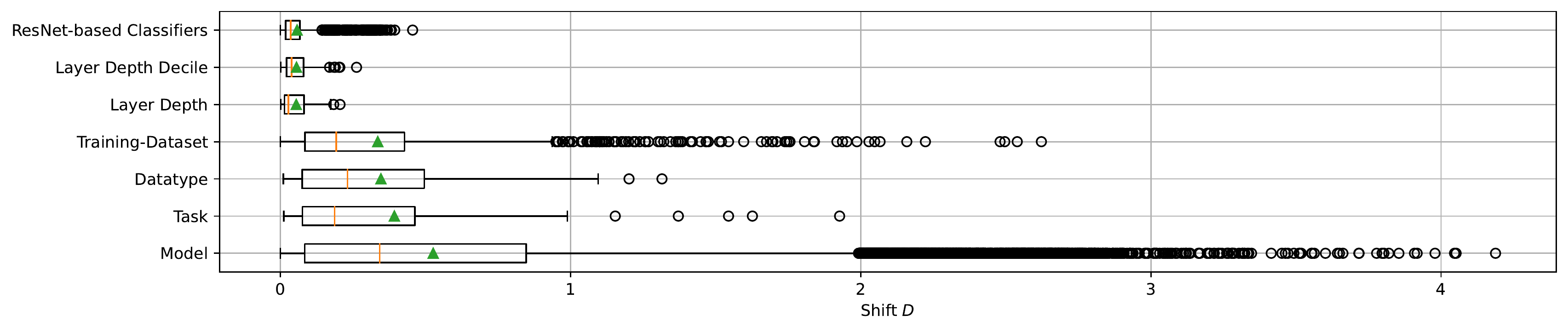}
  \caption{Distribution of the pair-wise shift $D$ for different filter sub-sets.}
  \label{fig:kl_boxplot}
\end{figure}

\vspace{-0.3cm}
\textbf{Between Models.} The comparison of filters between individual neural architectures results in the highest shift we observe for all investigated meta-groups (\autoref{fig:kl_boxplot}).

\textbf{Inside Model families.} The shift between models of the same family trained for the same task (e.g. \textit{ResNet}-classifiers in \autoref{fig:kl_boxplot}) is negligible and mostly independent of the training data, showing that the common practice of pre-training models with \textit{ImageNet} is indeed a valid approach even for visually distant application domains. 

\begin{figure}[H]
  \centering
  \includegraphics[width=\linewidth]{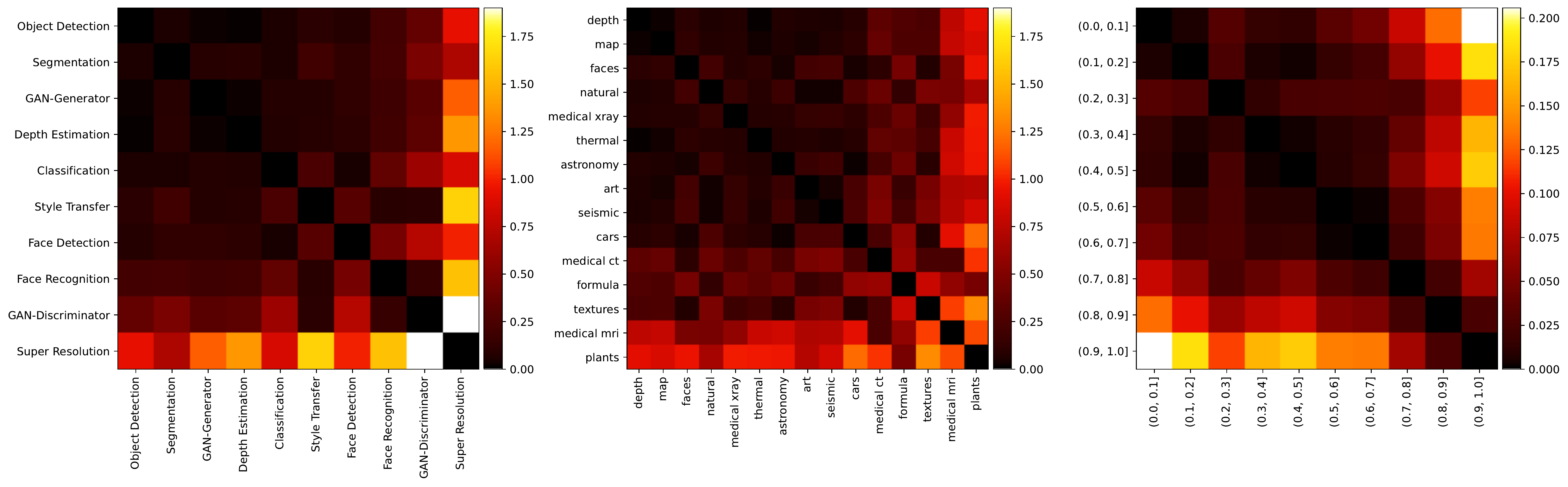}
  \begin{tabularx}{\linewidth}{>{\centering\arraybackslash}X>{\centering\arraybackslash}X>{\centering\arraybackslash}X}
        (a) & (b) & (c)\\
  \end{tabularx}
  \caption{$D$ matrices for different filter meta-groups: (a) by tasks, (b) by data types, (c) by layer depth decile (relative to the model depth).}
  \label{fig:kl_combined}
\end{figure}

\textbf{Between Tasks.} Unsurprisingly, classification, segmentation, object detection, and GAN-generator distributions are quite similar (\autoref{fig:kl_combined}{a}), since the non-classification models typically include a classification backbone. The smallest mean shift to other tasks is observed in object detection, GAN-generators, depth estimation models. 
Super Resolution models appears to be strong outliers, but we only have one model for this task. Additionally, this model contains \textit{PixelShuffle} layers \cite{shi2016realtime} that may tamper with the filters. Less transferable task distributions also include GAN-discriminators and face detection models. GAN-discriminators distributions do barely differ along principal components and can be approximated by a gaussian distribution. This indicates a filter distribution close to random initialization, representing a "confused" discriminator that cannot distinguish between real and fake samples towards the end of training.
\vspace{-0.5cm}
\vspace{-1cm}
\textbf{Data types and training sets.} We find that the distribution shift is well balanced across most data types and training sets (\autoref{fig:kl_combined}{b}). Most coefficient distributions tend to shrink towards the least significant principal components, with a remarkably wide distribution of the first principal component. Notable outliers include \textit{medical CT \& MRI, formula, texture,} and \textit{plants} data types. \textit{Medical} types have visible spikes in the KDEs, indicating that many structurally similar filters exist. The outlier \textit{Formula} includes models trained on \textit{Fractal-DB}, which was proposed as a synthetic pre-training alternative to \textit{ImageNet1k} \cite{KataokaACCV2020}.

\textbf{Layer depth.} The shift between layers of various depth deciles increases with the difference in depth (\autoref{fig:kl_combined}{c}), yet it is marginal compared to the shift across tasks or data types. Distributions in the last decile of depth form the most distinct interval, significantly outdistancing the first and second-to-last decile that follow next. However, splitting the coefficients by absolute depth introduces many extreme outliers (\autoref{fig:kl_boxplot}). That may again be a result of degenerated layers or an effect of under-sampling.

\vspace{-0.3cm}

\begin{minipage}{0.6\textwidth}
\subsubsection{Observable filter distribution phenotypes}
We categorize the distributions into three phenotypes depending on their distribution characteristic in the PCA space: I) distributions where all dimensions are gaussian-like; II) distributions containing one or more layer that shows a fairly small variety of feature patterns and therefore takes on discrete stages in bi-variate scatter plots; III) distributions where one or more distribution is multi-modal, not centered, highly sparse or otherwise non-normal forming scatter plots that look like symbols. Figure \ref{fig:scatter_pheno} shows typical examples of these phenotypes.
\end{minipage}
\hspace{0.01\textwidth}
\begin{minipage}{0.35\textwidth}
\begin{figure}[H]
    \centering
    \includegraphics[width=0.32\linewidth]{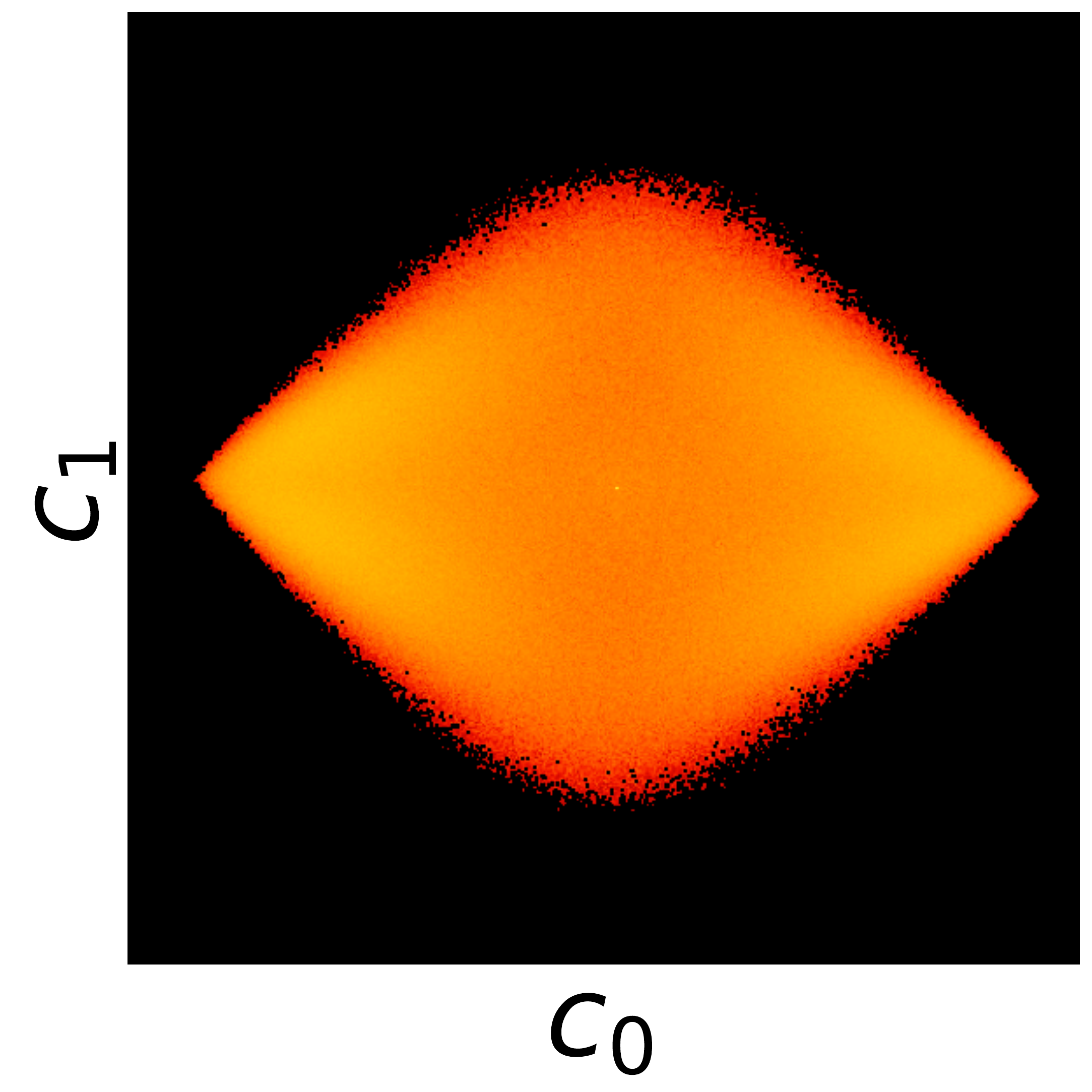}  
    \includegraphics[width=0.32\linewidth]{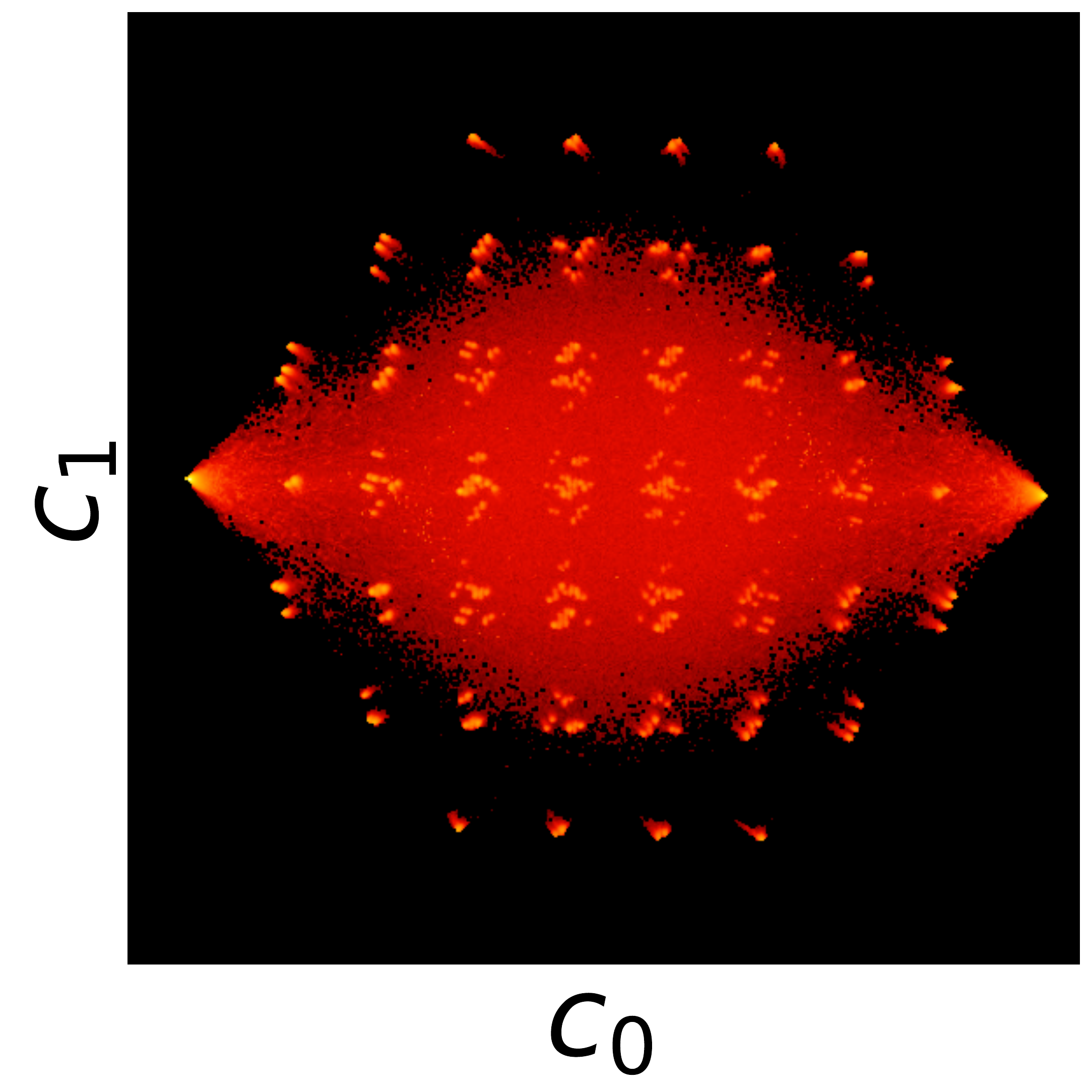}
    \includegraphics[width=0.32\linewidth]{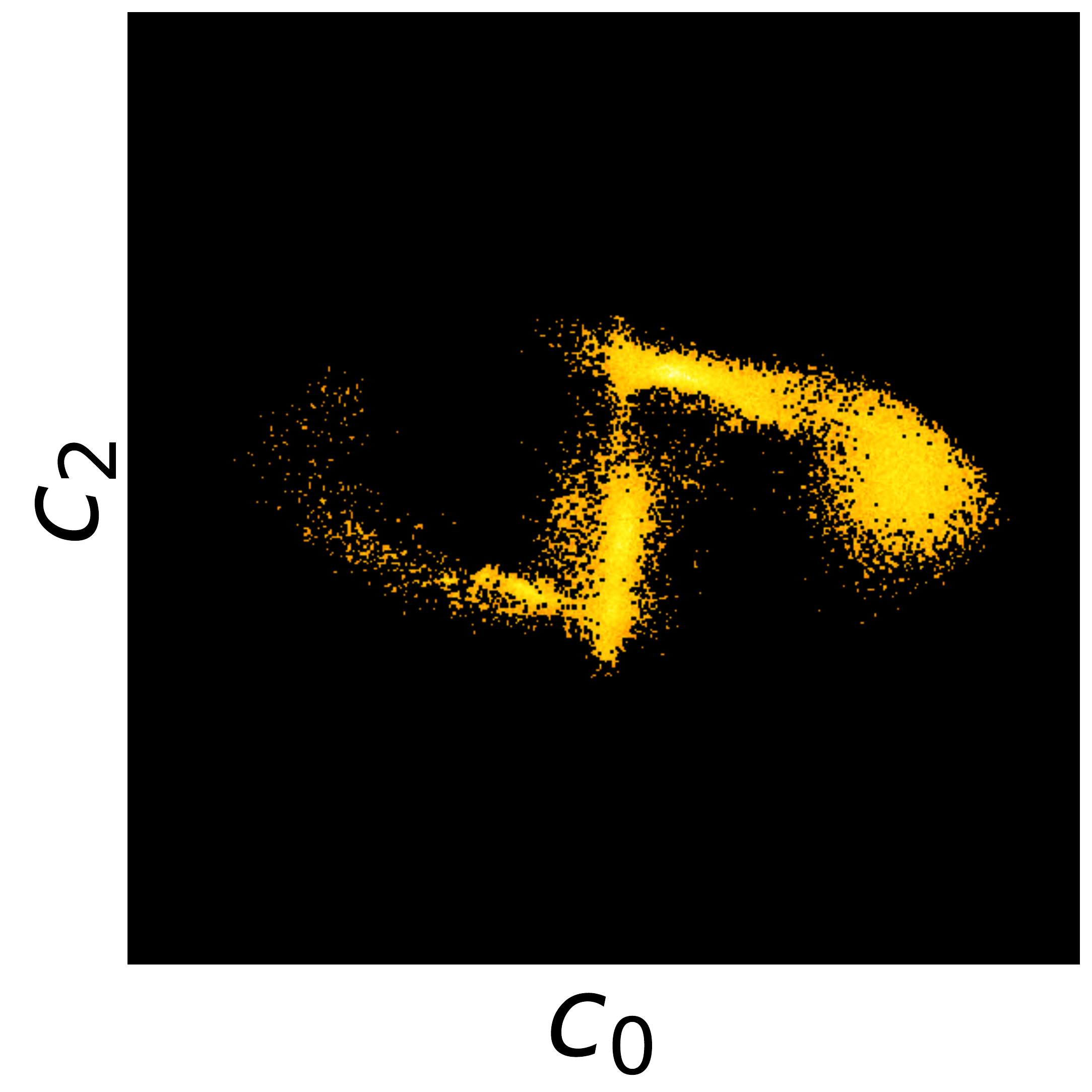}\\
    
    \begin{tabularx}{\textwidth}{>{\centering\arraybackslash}X>{\centering\arraybackslash}X>{\centering\arraybackslash}X}
        (a) & (b) & (c)\\
  \end{tabularx}

    \caption{Bi-variate plot between component distributions categorized as (a) sun, (b) spikes, (c) symbols.}
    \label{fig:scatter_pheno}
\end{figure}
\end{minipage}

\vspace{-0.3cm}
\subsection{Filter Scales}
\begin{figure}[H]
  \centering
  \includegraphics[width=\linewidth]{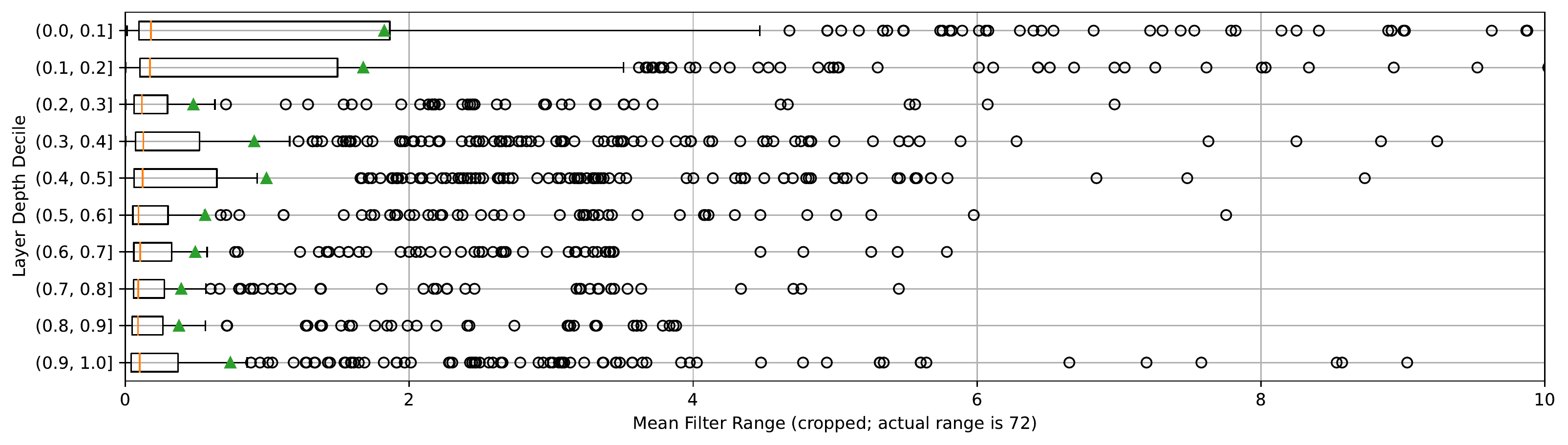}
  \caption{Boxplots showing mean range per layer depth decile (top to bottom in decreasing order) for each model in the dataset.}
  \label{fig:range_boxplot}
\end{figure}

\vspace{-0.3cm}
So far we have only studied the structural similarity, independent of the actual scale of the learned filter weights (difference between minimal and maximal weight). In \autoref{fig:range_boxplot} we compute the mean scale per layer depth decile. The distributions show an expected decrease with depth but also a high variance and many outliers across models, especially in the first two deciles.

\vspace{-0.3cm}
\section{Related Work}
\vspace{-0.3cm}

An extensive analysis of features, connections, and their organization extracted from trained \textit{InceptionV1} \cite{szegedy2014going} was presented in \cite{Olah2020,olah2020an,cammarata2020curve,olah2020naturally,schubert2021high-low,cammarata2021curve,voss2021visualizing,voss2021branch,petrov2021weight}. The authors of \cite{Yosinski2014} studied learned filter representations in \textit{ImageNet} classification models and presented the first moves towards transfer learning. A summary of transfer learning for image classification CNN can be found in \cite{hussain2019} and general surveys for other tasks and domains are available in \cite{5288526,zhuang2020comprehensive}. \cite{Aygun_2017_ICCV} captured convolution filter patterns with Gaussian Mixture Models to achieve cross-architecture transfer learning and \cite{tayyab2019basisconv} demonstrated that convolutions filters can be replaced by a fixed filter basis that $1\times 1$ convolution layers blend.\\
A benchmark for distribution shifts that arise in real-world applications is provided in \cite{pmlr-v139-koh21a}  and \cite{taori2020measuring} measured robustness to natural distribution shifts.
Lastly, \cite{Djolonga_2021_CVPR} studied the correlation between transfer performance and distribution shifts of image classification models. 

\vspace{-0.3cm}
\section{Discussion and Outlook}
\vspace{-0.3cm}
Our first results support our initial hypothesis that the distributions of trained convolutional filters are a suitable and easy-to-access proxy for the investigation of image distributions and the similarity between the same. While the presented results are still in the early stages of a thorough study, we report several interesting findings that could be explored to obtain better model generalizations and assist in finding suitable pre-trained models.   
One finding is the presence of large amounts of degenerated (or untrained) filters in large, well-performing networks - resulting in the phenotypes \textit{spikes} and \textit{symbols}. We assume that their existence is a symptom in line with the \textit{Lottery Ticket Hypothesis} \cite{frankle2018lottery}.\\
Another striking finding is the observation of very low shifts between different meta-groups: I) shifts inside a family of architectures are very low, independent of the target image distribution; II) also we observe rather small shifts between convolution layers of different depths;  while III) shifts between different tasks are higher, even when related architectures and the same data is used.        

\newpage

\begin{ack} 
This work was supported in part by the German Ministry for Science, Research and Arts Baden-Wuerttemberg (MWK) under Grant 32-7545.20/45/1 Quality Assurance of Machine Learning Applications (Q-AMeLiA). \url{https://q-amelia.in.hs-furtwangen.de/}.
\end{ack}

\bibliographystyle{unsrt}
\bibliography{main}

\appendix

\newpage
\section{Appendix}

\subsection{Detailed divergence computation}

All probability distributions are represented by histograms. The histogram range is defined by the minimum and maximum value found across all distributions to compare. 70 uniform bins are used.

\subsection{D-matrix for non-scaled filters}
\begin{figure}[H]
  \centering
  \includegraphics[width=\linewidth]{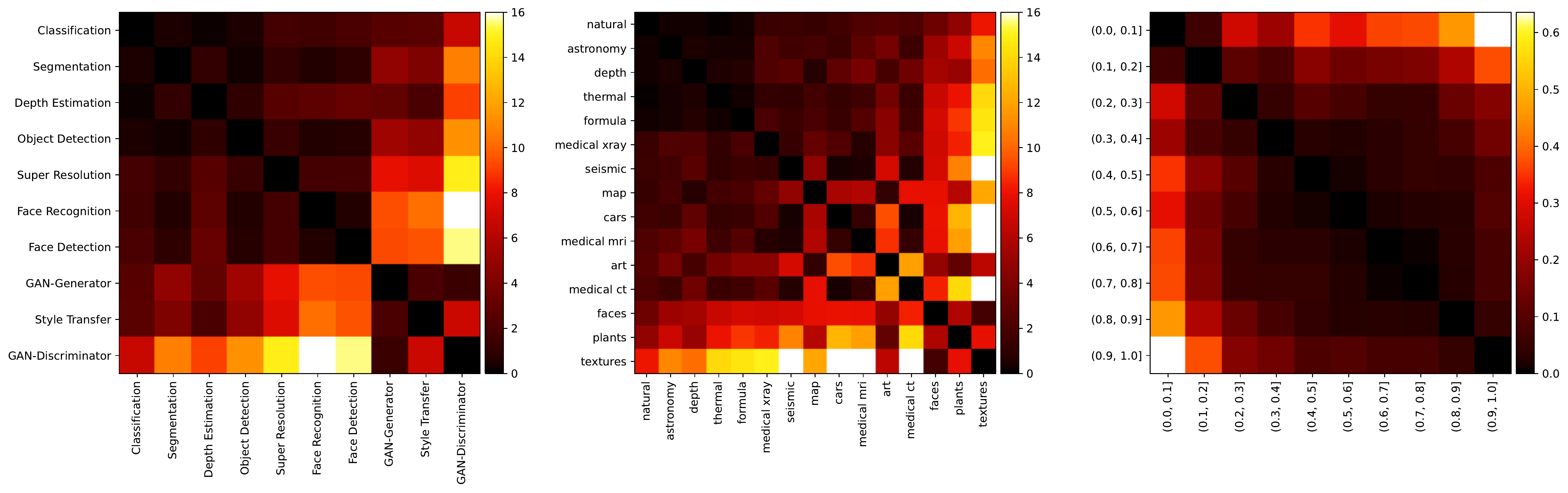}
  \begin{tabularx}{\linewidth}{>{\centering\arraybackslash}X>{\centering\arraybackslash}X>{\centering\arraybackslash}X}
        (a) & (b) & (c)\\
  \end{tabularx}
  \caption{$D$ matrices for different filter sub-sets on the raw filter data (non-scaled): (a) tasks, (b) data types, (c) filter depth decile relative to the model depth.}
  \label{fig:appendix_kl_combined_raw}
\end{figure}

\subsection{PCA}\label{sec:appendix_pca}

\begin{figure}[H]
    \centering
    \begin{tabular}{m{0.45\linewidth} m{0.45\linewidth}}
    (a) \adjustbox{valign=c}{\includegraphics[width=0.925\linewidth]{figures/static/basis_all.pdf}} &   
    (b) \adjustbox{valign=c}{\includegraphics[width=0.925\linewidth]{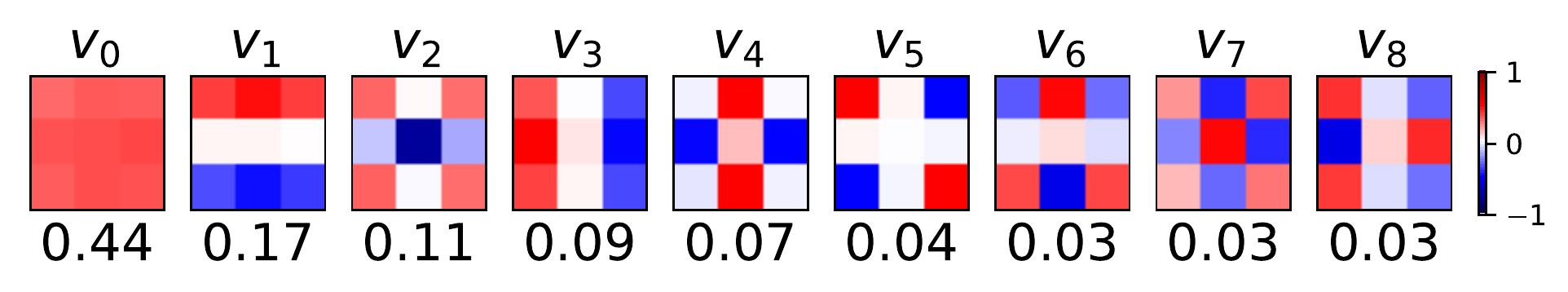}} \\
    (c) \adjustbox{valign=c}{\includegraphics[width=0.925\linewidth]{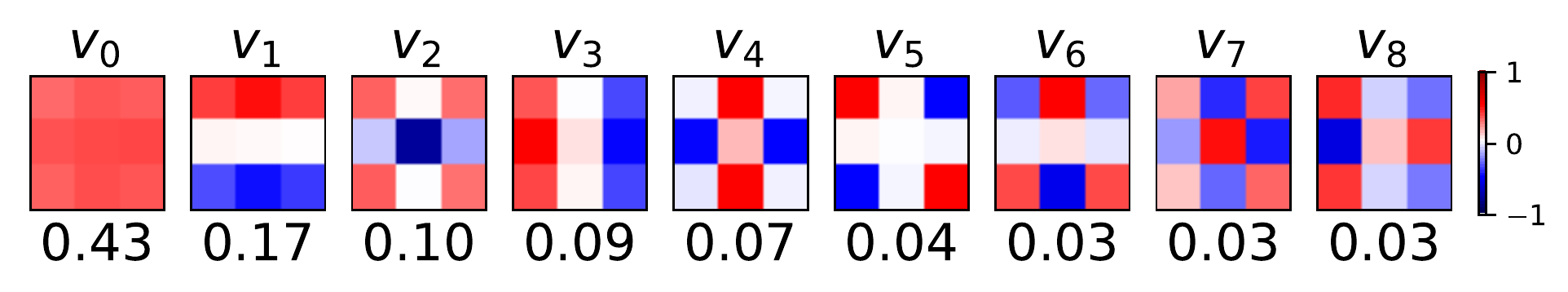}} & 
    (d) \adjustbox{valign=c}{\includegraphics[width=0.925\linewidth]{figures/static/basis_formula.pdf}} \\
    (e) \adjustbox{valign=c}{\includegraphics[width=0.925\linewidth]{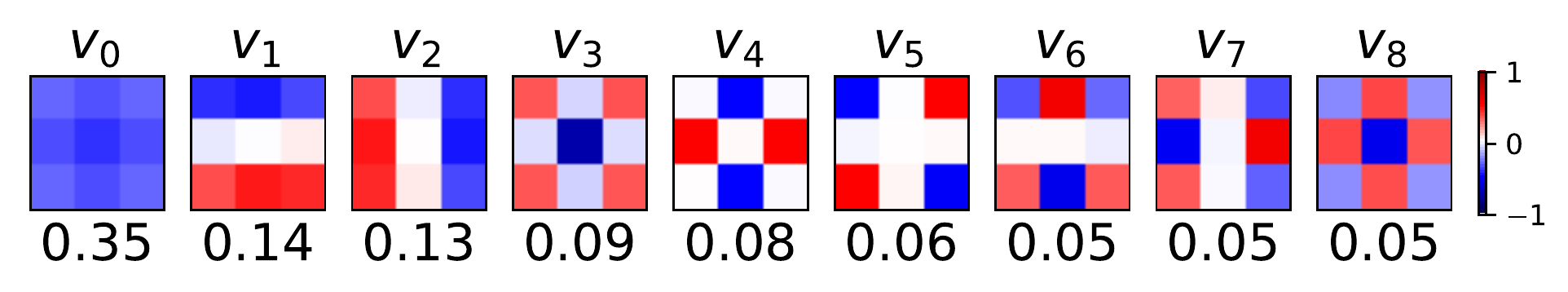}} &
    (f) \adjustbox{valign=c}{\includegraphics[width=0.925\linewidth]{figures/static/basis_gan_disc.pdf}} \\
    (g) \adjustbox{valign=c}{\includegraphics[width=0.925\linewidth]{figures/static/basis_first_layers.pdf}} &
    (h) \adjustbox{valign=c}{\includegraphics[width=0.925\linewidth]{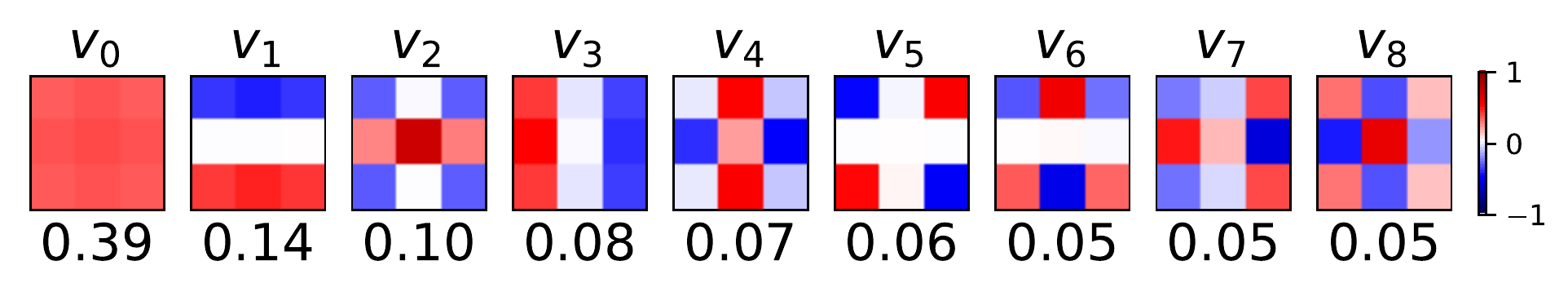}} \\
    \end{tabular}
    \caption{More principal components $v_i$ and explained variance ratio per component for filters from (a) all, (b) classification models, (c) models trained on \textit{ImageNet1k}, (d) models trained on formula data, (e) GAN-generators, (f) GAN-discriminators, (g) first convolution layers, (h) last convolution layers. Computed on the scaled data set.}
    \label{fig:appendix_filter_basis}
\end{figure}

\begin{figure}[H]
    \centering
    \begin{tabularx}{\linewidth}{>{\centering\arraybackslash}X>{\centering\arraybackslash}X}
    \includegraphics[width=0.9\linewidth]{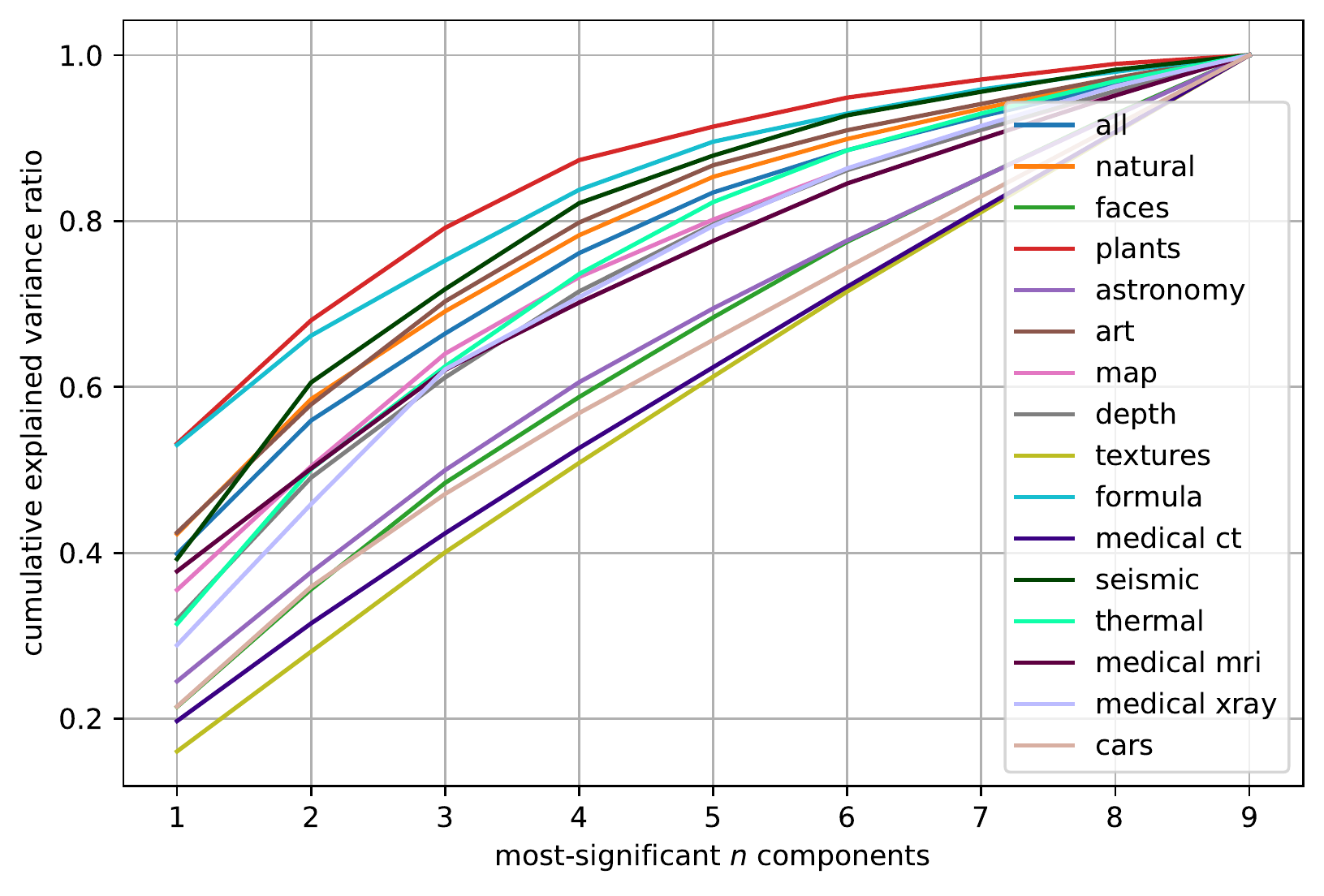} &   
    \includegraphics[width=0.9\linewidth]{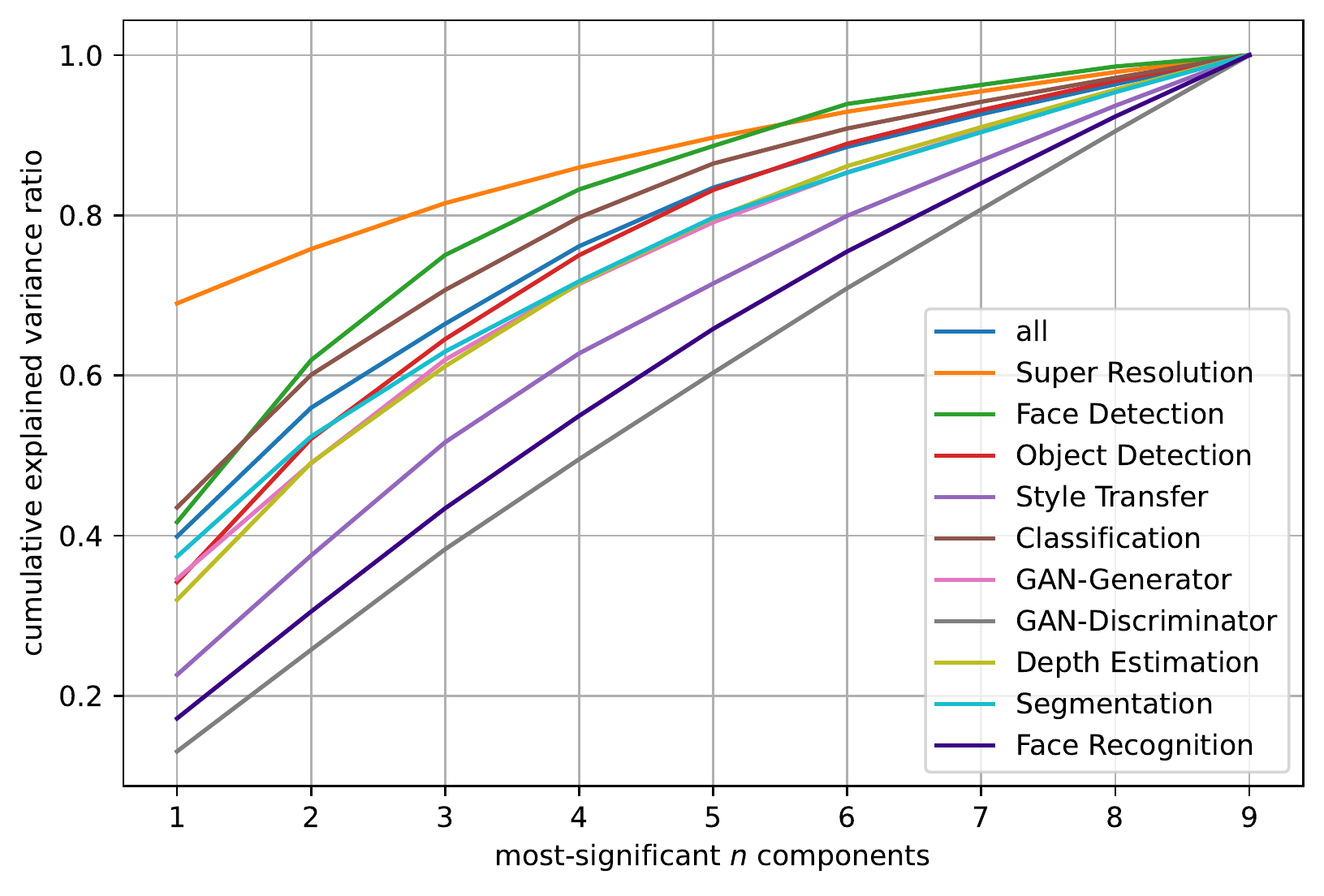} \\
    (a) data type & (b) task\\
    \end{tabularx}
    \caption{Cumulative ratio of explained variance over the first $n$ components by all tasks and data types.}
    \label{fig:pca_cumsum}
\end{figure}

\subsection{Data set statistics}\label{sec:appendix_stats}

\begin{figure}[H]
  \centering
  \includegraphics[width=\linewidth]{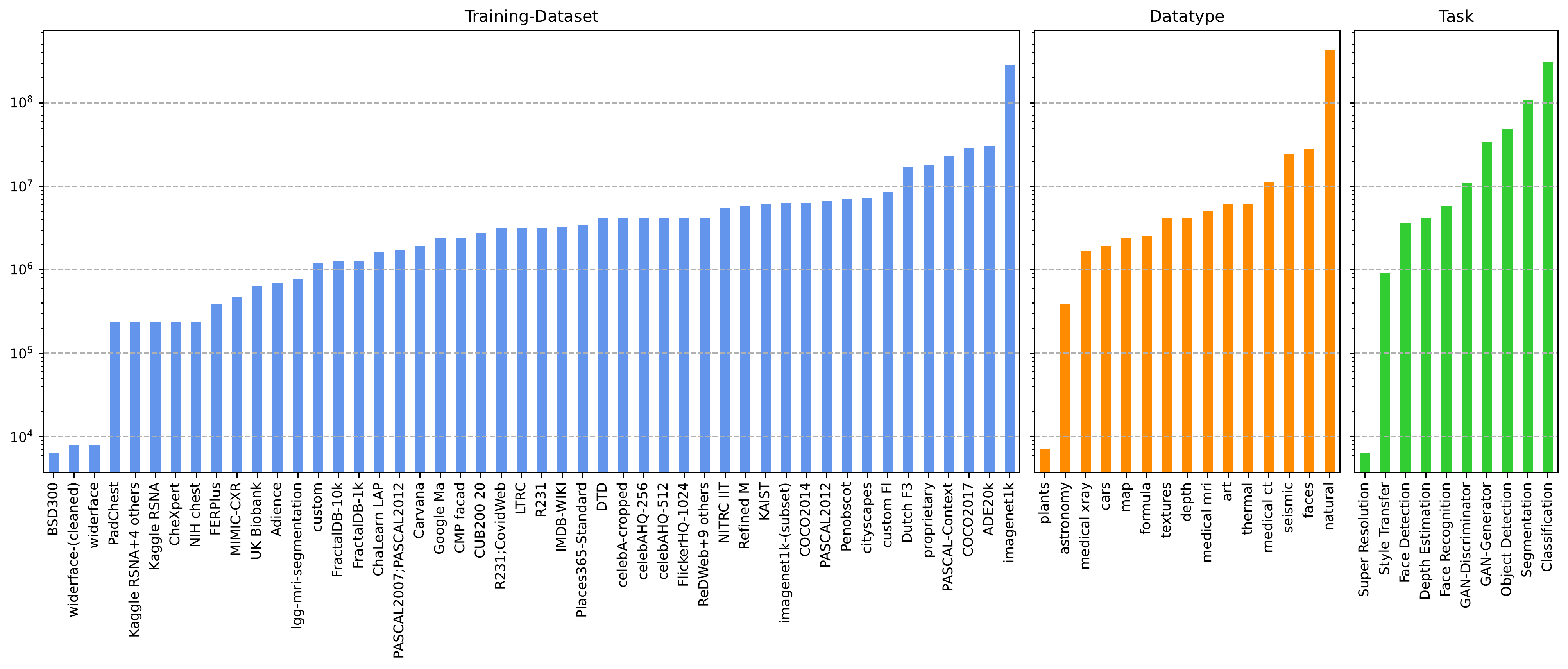}
  \caption{Total count of filters per filter sub-set. Log scale.}
  \label{fig:stats_filters_by_split}
\end{figure}

\begin{figure}[H]
  \centering
  \includegraphics[width=\linewidth]{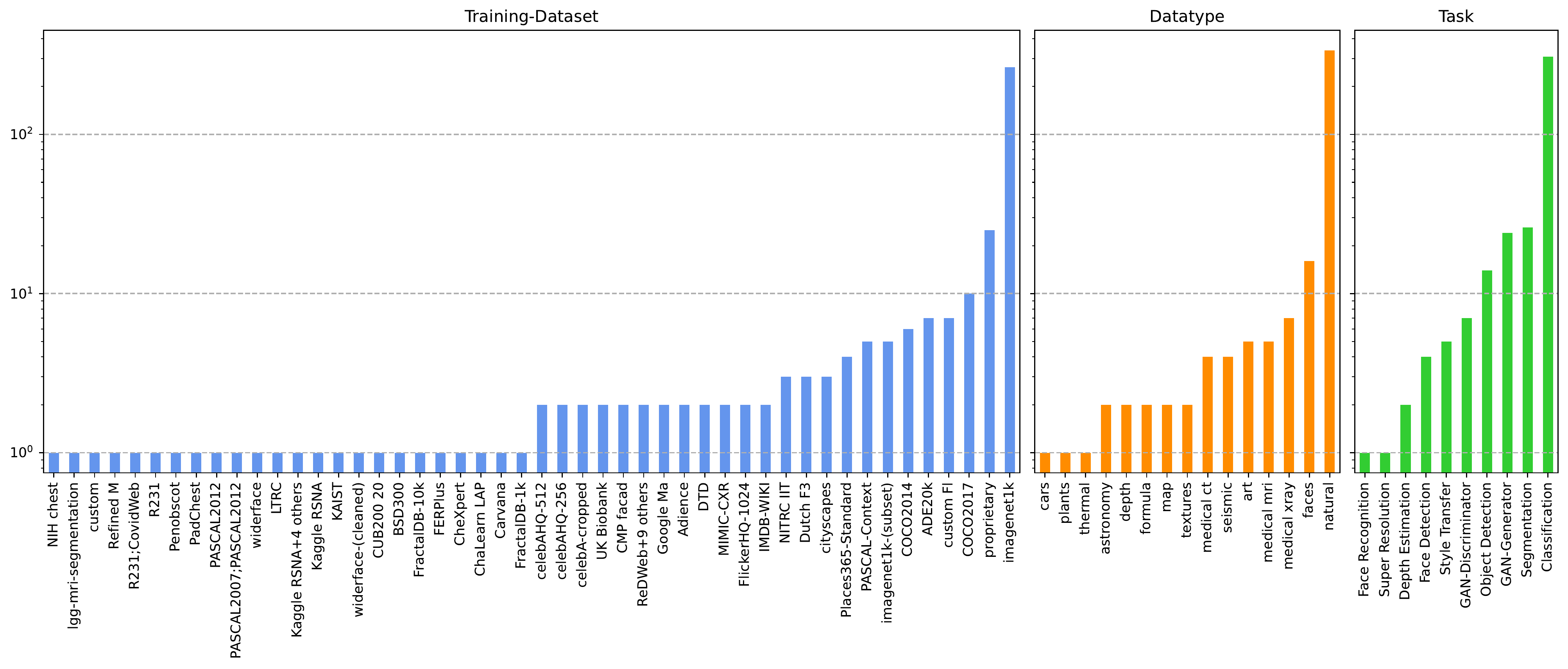}
  \caption{Total count of models per filter sub-set. Log scale.}
    \label{fig:stats_models_by_split}
\end{figure}

\subsection{Ridge Plots}\label{sec:appendix_ridge_plots}

\begin{figure}[H]
  \centering
  \includegraphics[width=\linewidth]{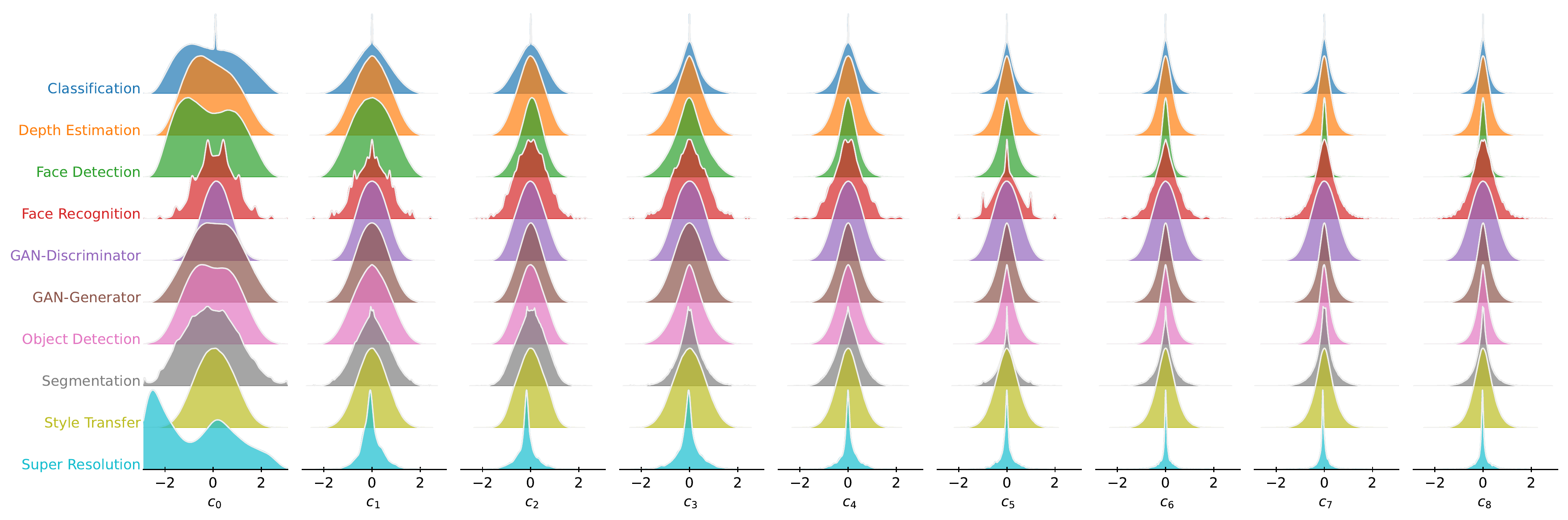}
  \caption{Distribution of the coefficients along the principal components by\textbf{ model task}.}
    \label{fig:ridge_task}
\end{figure}

\begin{figure}[H]
  \centering
  \includegraphics[width=\linewidth]{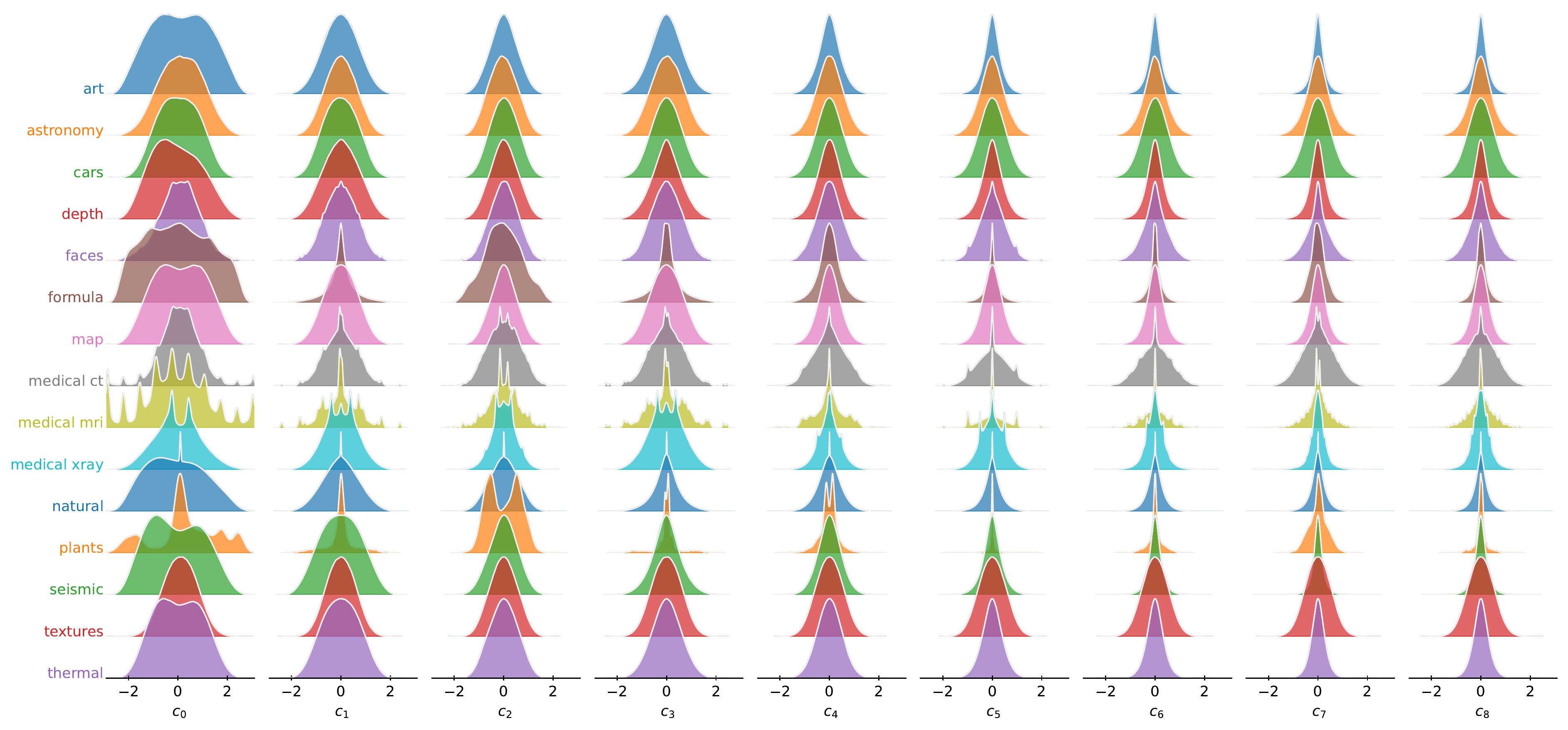}
  \caption{Distribution of the coefficients along the principal components by training \textbf{data type}.}
  \label{fig:ridge_datatype}
\end{figure}

\begin{figure}[H]
  \centering
  \includegraphics[width=\linewidth]{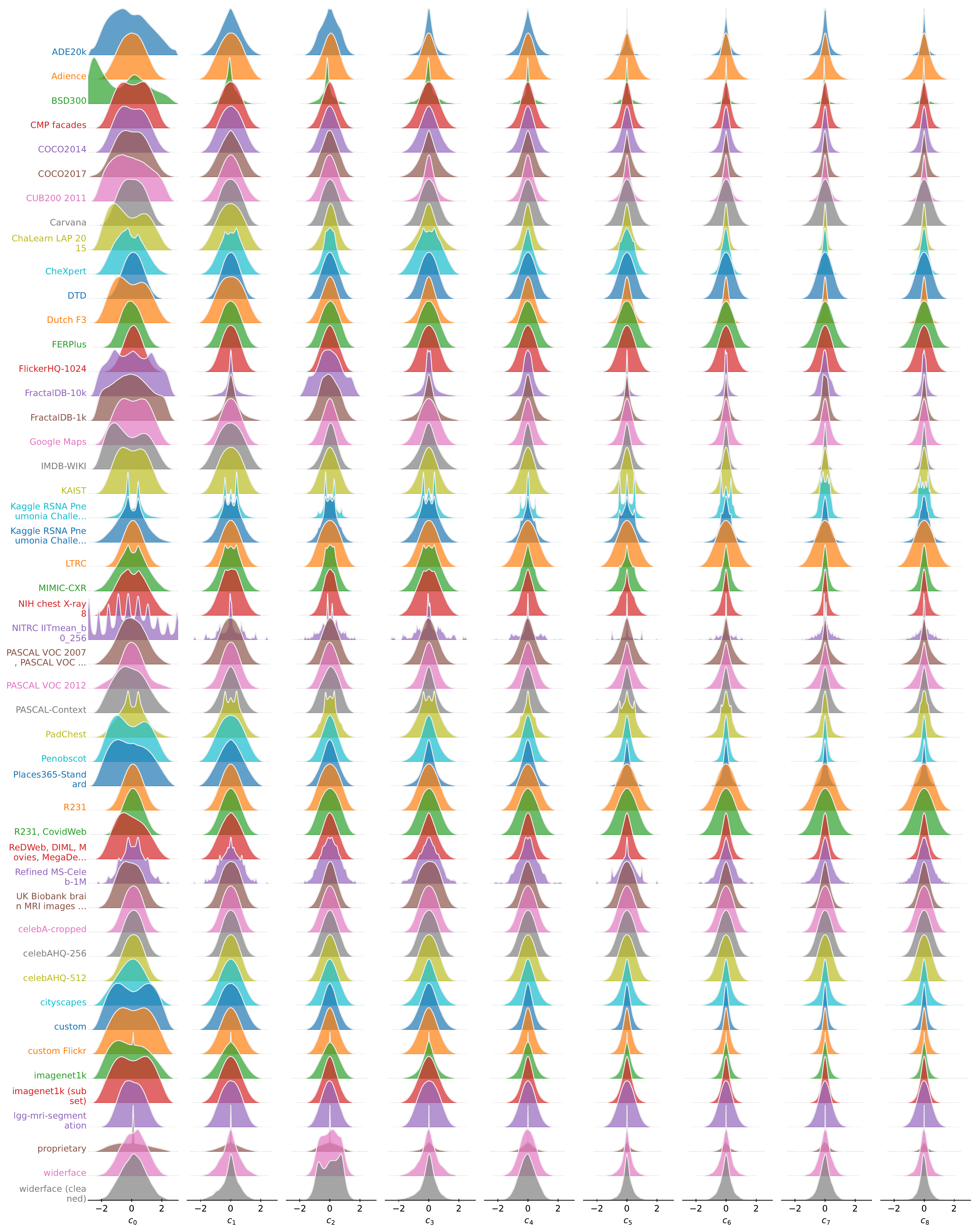}
  \caption{Distribution of the coefficients along the principal components by combination of \textbf{data set} used for training.}    
  \label{fig:ridge_training}
\end{figure}

\begin{figure}[H]
  \centering
  \includegraphics[width=\linewidth]{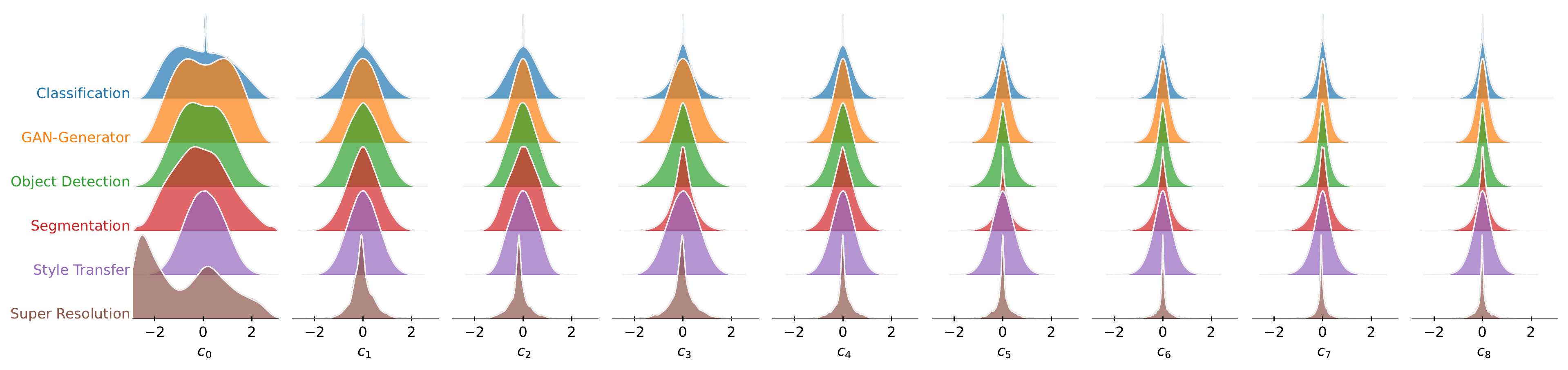}
  \caption{Distribution of the coefficients along the principal components by model \textbf{task for} data sets labeled as \textbf{natural data type}.}
  \label{fig:ridge_task_for_natural}
\end{figure}

\begin{figure}[H]
  \centering
  \includegraphics[width=\linewidth]{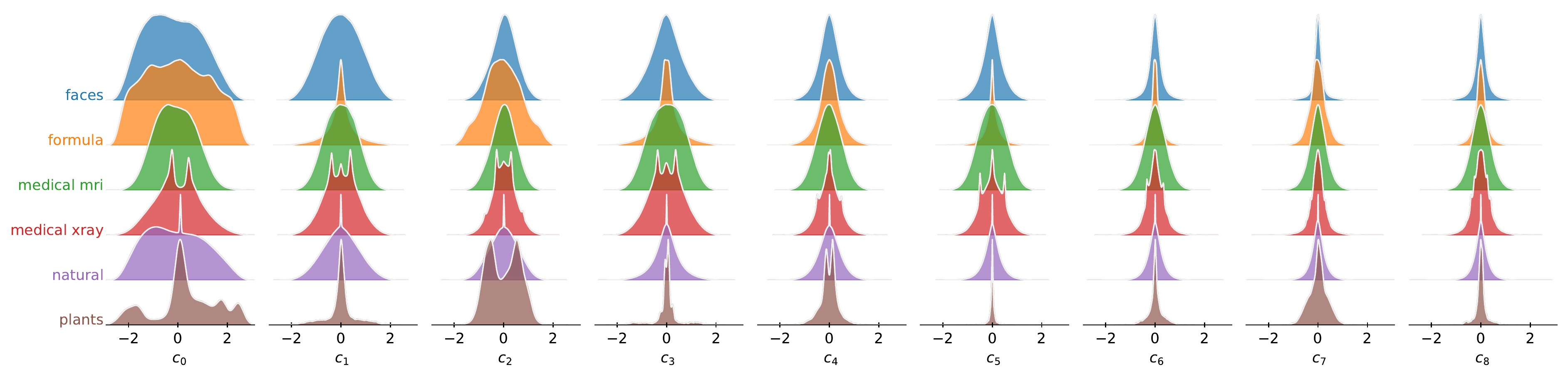}
  \caption{Distribution of the coefficients along the principal components by training \textbf{data type for image classification models.}}
  \label{fig:ridge_datatype_for_classification}
\end{figure}

\begin{figure}[H]
  \centering
  \includegraphics[width=\linewidth]{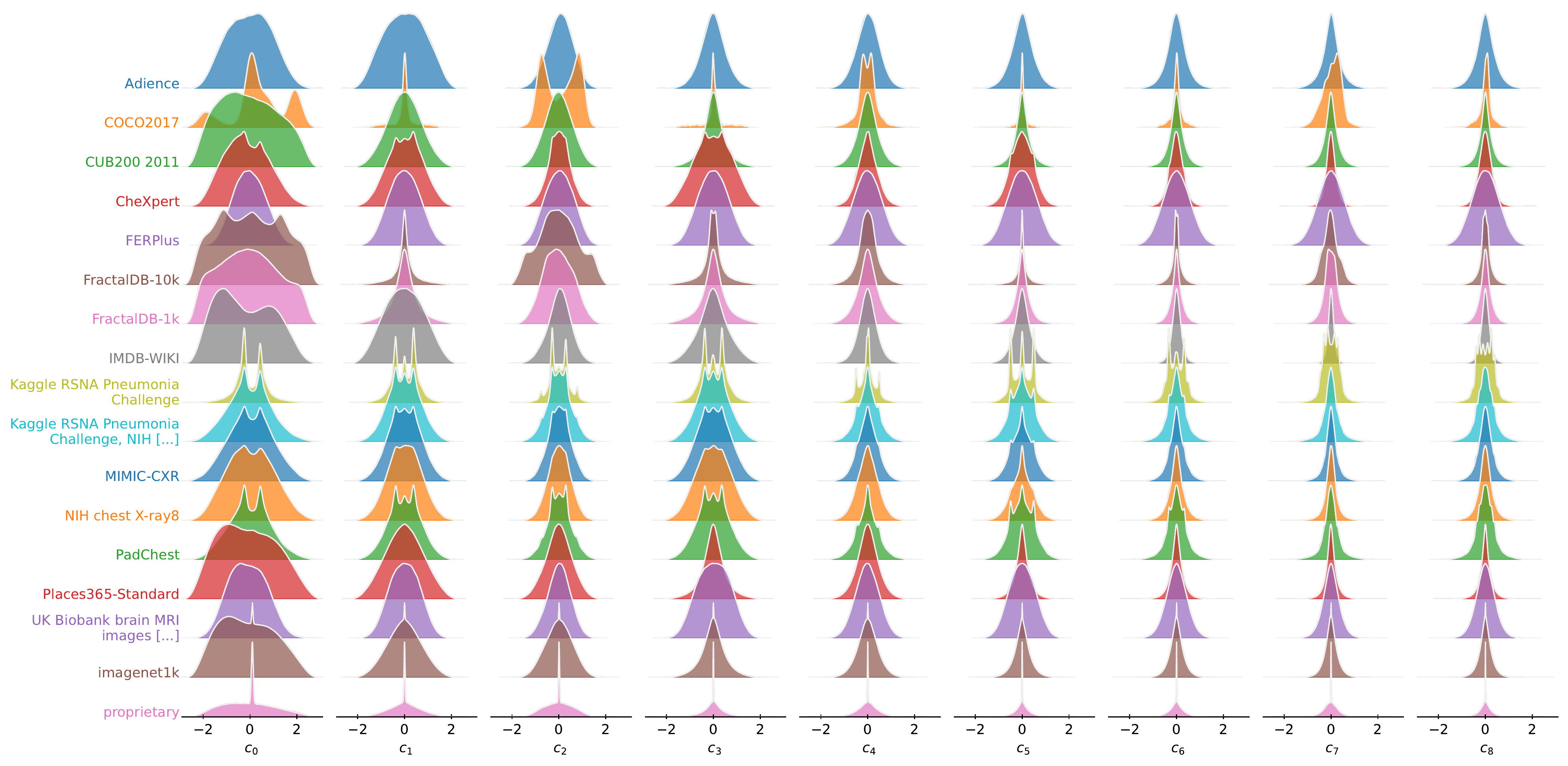}
  \caption{Distribution of the coefficients along the principal components by training \textbf{data set for image classification} models.}
  \label{fig:ridge_training_for_classification}
\end{figure}

\end{document}